# Pars-ABSA: an Aspect-based Sentiment Analysis dataset for Persian

**Taha Shangipour ataei, Kamyar Darvishi, Soroush Javdan, Behrouz Minaei-Bidgoli, Sauleh Eetemadi**

Department of Computer Engineering, Iran University of Science and Technology, Tehran, Iran

{taha_shangipour, kamyar_darvishi, soroush_javdan}@comp.iust.ac.ir

{b_minaei, sauleh}@iust.ac.ir

## Abstract

Due to the increased availability of online reviews, sentiment analysis had been witnessed a booming interest from the researchers. Sentiment analysis is a computational treatment of sentiment used to extract and understand the opinions of authors. While many systems were built to predict the sentiment of a document or a sentence, many others provide the necessary detail on various aspects of the entity (i.e. aspect-based sentiment analysis). Most of the available data resources were tailored to English and the other popular European languages. Although Persian is a language with more than 110 million speakers, to the best of our knowledge, there is a lack of public dataset on aspect-based sentiment analysis for Persian. This paper provides a manually annotated Persian dataset, Pars-ABSA, which is verified by 3 native Persian speakers. The dataset consists of 5,114 positive, 3,061 negative and 1,827 neutral data samples from 5,602 unique reviews. Moreover, as a baseline, this paper reports the performance of some state-of-the-art aspect-based sentiment analysis methods with a focus on deep learning, on Pars-ABSA. The obtained results are impressive compared to similar English state-of-the-art.

## 1 Introduction

Nowadays, with the rapid growth of the Internet, a huge volume of data is generated daily. And it is said that the world's most valuable resource is no longer oil but data. On the other hand, humans are always curious about how others think and want to use others' recommendation. User opinion plays a vital role in the modern world, for instance, manufacturing companies use online reviews to get a sense of general sentiment for their products. Also, they study user reactions and reply to them on microblogs and use this media as an opportunity to market their products. As well as, taking advice from others' opinions is an important part of the decision making procedure. According to two surveys of more than 2000 American adults, 81% of them have done online research on a product at least once; and between 78% and 87% of readers of online reviews of various services (like restaurants, hotels and etc.) reported that reviews had a great impact on their purchase (Pang et. al., 2008). Among services available over the internet which mentions: personal blogs, social media, recommendation services and also e-commerce websites as the main sources of reviews and user opinions which allow people to share and express their views about topics, discuss about current issues and express positive or negative sentiment toward products they use in daily life with others around the world.

Personal blog: people share their routines activities through their blogs. An influencer can easily persuade people to buy a product by telling others about his experience of using it.

Social media: another medium for sharing opinions is social media. People use services such as Twitter, Instagram and Facebook, which are a place to post real-time short messages.

E-Commerce: websites like Amazon and E-bay are another source of user opinion. People who buy products share their experience of using products over these websites with others.

By being at the era of data explosion, for example around 500 million tweets are sent daily, one challenge is to build a system to detect and summarize an overall sentiment of these data.

Sentiment analysis is the computational study of detecting and extracting subjective information and attitudes about entities. The entity can represent individuals, events, products or topics. The output of it is opinion polarity. Polarity is generally expressed in different forms from two classes of positive and negative or three classes of positive, neutral and negative while at some researches it is represented as a real number between 1-5 stars or 0-10 grade. Sentiment analysis was acknowledged at the early 2000s with (Turney, 2002) and (Pang et al., 2002), both of them doing binary classification on reviews with two classes of positive and negative. At (Turney, 2002), an unsupervised learning algorithm is used for classifying the reviews and the dataset used is from four different domains which consist of automobiles, banks, movies, and travel destinations. At (Pang et al., 2002), they used supervised machine learning methods and the dataset is collected from IMDB movie reviews. At (Pang and Lee, 2005), they changed polarity from binary (positive and negative) to a multi-point scale (one to five stars). Data sources used for models are generally collected from tweets, blogs and, reviews about movies and products like (Pang and Lee, 2005) and (Branavan et al., 2009). Sentiment analysis can also be useful for politics. For instance at presidential elections, candidates can understand public opinions and establish their policies by the use of sentiment analysis approaches. At (Ramteke et al., 2016), a machine learning model is provided to predict the results of the 2016 United States presidential election based on collected tweets. Sentiment analysis is generally performed at three different levels: document-based, sentence-based, and aspect-based.

At document-based, it is considered that the whole document, for instance, in a movie review is the movie is an entity and the whole document expresses a positive, negative or neutral polarity about it. All of the preceding reviewed articles were at document-based. There are many datasets available for this task from product review to tweets and hotel reviews, but two of the well-studied datasets are discussed in (Maas et al., 2011) which consists of 50,000 reviews from IMDB website and Yelp reviews (Zhang et al., 2015) with more than 500,000 data samples in both binary and five-class versions.

At sentence-based, the goal is to predict and determine the polarity of each sentence, in other words, each sentence at a document is separated from others and may have a different polarity. In Stanford sentiment tree-bank (Socher et al., 2013), the task of fine-grained classification is to determine one of five labels (very negative, negative, neutral, positive, very positive) to a sentence from a document. Most of the available sentiment analysis tools are concerned with detecting the polarity of a sentence or a document. Aspect-based sentiment analysis is the latest task in this field which is focused on identifying the polarity of the targets expressed in a sentence. A target is an object, and its components, attributes, and features. For instance, at (Liu, 2010) a model is provided that identifies the polarity of product features that have been commented on by the reviewer. As an example, in "Food was great but the service was miserable." There are two opinion targets, "food" and "service". The reviewer has a positive sentiment polarity on "food" and a negative sentiment polarity on "service". This example shows why document-based and sentence-based systems are not sufficient for this task. The superiority of aspect-based models to sentence-based and document-based models becomes vivid when manufacturers or service providers want to know which component or feature of their product is not well enough and needs to improve based on the negative reviews they get from customers.

The rest of the paper is organized as follows. In section 2, sentiment analysis datasets available for Persian and English are discussed. In section 3, details about the data collection and annotation are given. In section 4 result of applying state-of-the-art systems available for aspect-based sentiment analysis on Pars-ABSA dataset is provided and discussed. In section 5 we conclude and give future directions of research.

## 2 Related Works

Persian is the official language of Iran, Afghanistan, and Tajikistan, and also is spoken in Uzbekistan with more than 110 million speakers. Persian is generally classified as western
Iranian languages and is from the Indo-European family. In document-based sentiment analysis for Persian, (Saraee and Bagheri, 2013) and (Bagheri et al., 2013) provided a dataset of manually annotated reviews of cellphones from 829 online reviews; and, at (Hajmohammadi and Ibrahim, 2013), a dataset of 400 manually annotated from Persian movie reviews was proposed. For sentence-based, at (Basiri and Kabiri, 2017) proposed SPerSent, a Persian dataset consisting of 150,000 sentences from product reviews of Digikala website. Each sentence is associated with two types of labels, binary (Positive and Negative) and five-star rating and is labeled automatically based on the majority voting of three different lexicons.

| Dataset | | Positive | Neutral | Negative |
|---|---|---|---|---|
| Laptop | Train | 980 | 454 | 858 |
| | Test | 340 | 171 | 128 |
| Restaurant | Train | 2159 | 632 | 800 |
| | Test | 730 | 196 | 195 |
| Twitter | Train | 1567 | 3127 | 1563 |
| | Test | 174 | 346 | 174 |

Table 1 – Number of data samples for each sentiment polarity of 3 English datasets

In last decade, in aspect-based sentiment analysis, most of the data resources and systems built so far are tailored to English and other languages like Chinese and Arabic. There are three datasets for English which are mainly used by researchers to compare the performance of their systems. These three datasets are Restaurants and Laptops (Pontiki et al., 2014) and Twitter (Dong et al., 2014). First and second datasets are annotated data samples from comments and reviews about laptops and restaurants from Semeval-2014 task 4: Aspect-based sentiment analysis and the last one is collected tweets from Twitter. The number of data samples in each dataset is given in Table 1.

At (Saeidi et al., 2016), Sentihood is presented, which consists of annotated data from a question answering platform in the domain of neighborhoods of a city. Along with this dataset, the task of targeted aspect-based sentiment analysis is introduced which is different from general aspect-based sentiment analysis, in extracting fine-grained information with respect to targets specified in reviews. At (Chen et al., 2017), a Chinese dataset for aspect-based sentiment analysis from comments about the news with 6365 positive, 9457 neutral and 6839 negative annotated data samples was proposed.

To the best of our knowledge, there is no available research on aspect-based sentiment analysis for Persian which is due to the lack of publically available datasets for this language. It is noteworthy to mention that until now, SentiPers (Hosseini et al., 2018) corpus was the only published dataset that contains annotated data in all three levels (document-based, sentence-based, and aspect-based) with 26,767 sentences and 21,375 target words from product reviews. The number of sentences and opinion words for each polarity category in SentiPers (Hosseini et al., 2018) is given at Table 2.

| | Polarity | | |
|---|---|---|---|
| | Positive | Neutral | Negative |
| Opinion words | 21,471 | 1,661 | 3,864 |
| Sentences | 12,921 | 11,353 | 2,678 |

Table 2 – Number of opinion words and sentences in SentiPers dataset

| Property | Value |
|---|---|
| Number of targets | 10,002 |
| Number of targets with positive polarity | 5,114 |
| Number of targets with negative polarity | 3,061 |
| Number of targets with neutral polarity | 1,827 |
| Number of tokens | 693,825 |
| Number of unique words | 18,270 |
| Number of comments | 5,602 |
| The average number of words per comments | 123.85 |

Table 3 – Properties of Pars-ABSA dataset

## 3  Data Collection and Annotation

This article provides the first totally aspect-based dataset for Persian. Pars-ABSA was created from collected user reviews from Digikala[1] website. Digikala is the biggest e-commerce startup in Iran and thousands of people buy goods from its website daily, and some of them post comments about the products they bought and share their experiences with others. More than 600,000 comments were scraped from Digikala website till January 2019. For manually annotating data samples a framework with Python language and Jupyter notebook was made. After the completion of annotation process, the dataset was confirmed and verified by 3 participants who were native

---

[1] http://www.digikala.com, Based on the terms of Digikala, the information of their website is allowed to be used for non-commercial activities with referring to them.

Persian speakers. Statistical information about the proposed dataset is given in Table 3.

```xml
<?xml version="1.0" encoding="UTF-8"?>
<sentences>
  <sentence id="102">
    <text>من اتفاقا بوی یاس احساس نمیکنم بیشتر بوی گلابی استشمام میکنم ولی با اینکه عطر متداولی نیست نمیدونم چرا بوش بنظرم تکراری اومد و انگار قبلا خیلی تجربش کرده بودم در کل خوش بو با پخش بوی کم و ماندگاریه متوسطه</text>
    <aspectTerms>
      <aspectTerm to="198" from="190" polarity="neutral" term="ماندگاری"/>
      <aspectTerm to="184" from="177" polarity="negative" term="پخش بوی"/>
    </aspectTerms>
  </sentence>
  <sentence>
    <text>در کل به نظرم اسپیکر خوش فرم و خوبی میاد جنس بدنه ی خوبی داره</text>
    <aspectTerms>
      <aspectTerm to="48" from="38" polarity="positive" term="جنس بدنه ی"/>
    </aspectTerms>
  </sentence>
</sentences>
```

Fig. 1 – An example of data samples stored in XML format

Pars-ABSA dataset is stored in two formats: 1-XML 2-Text. In XML format, there is a main tag named "sentences" that contains all of the data samples. For each review in dataset, there is a "sentence" tag available inside the main tag. "sentence" tag encompasses two types of tags, first is "text" tag that has the review text and the second is "aspectTerms" that consists of one or more "aspectTerm" tags. As long as it is possible to have more than one aspect in each sentence, all of the available aspects in each sentence are going to be inside "aspectTerms" and for each aspect, there would be an "aspectTerm" tag with four attributes. These attributes are:

- from: the starting position of aspect term in the sentence
- to: the ending position of aspect term in the sentence
- term: aspect term in the sentence
- polarity: sentiment polarity of aspect term

An example of stored data samples in XML format is given at Fig. 1

من اتفاقا بوی یاس احساس نمیکنم بیشتر بوی گلابی استشمام میکنم ولی با اینکه عطر متداولی نیست نمیدونم چرا بوش بنظرم تکراری اومد و انگار قبلا خیلی تجربش کرده بودم در کل خوش بو با پخش بوی کم و $T$ متوسطه
ماندگاری
.
من اتفاقا بوی یاس احساس نمیکنم بیشتر بوی گلابی استشمام میکنم ولی با اینکه عطر متداولی نیست نمیدونم چرا بوش بنظرم تکراری اومد و انگار قبلا خیلی تجربش کرده بودم در کل خوش بو با $T$ کم و ماندگاری متوسطه
پخش بوی
-1
در کل به نظرم اسپیکر خوش فرم و خوبی میاد $T$ خوبی داره
جنس بدنه ی
1

Fig. 2 – An example of data samples stored in Text format

In the second format, for each aspect term, there are three lines inside the file, the review is at first line but the aspect term is replaced with "$T$", aspect term is written in second line and in third line, there is a number available for sentiment polarity of the aspect term (1 for positive, 0 for neutral and -1 for negative). An example of data samples in the text format is available at Fig. 2

| Model | Twitter | | Laptop | | Restaurant | |
|---|---|---|---|---|---|---|
| | Acc | F1 | Acc | F1 | Acc | F1 |
| AOA | | | 74.5 | | **81.2** | |
| Cabasc | **71.53** | | **75.07** | | 80.89 | |
| RAM | 69.36 | **67.3** | 74.49 | **71.35** | 80.23 | **70.8** |
| IAN | | | 72.1 | | 78.6 | |
| TD-LSTM | 66.62 | 64.01 | 71.83 | 68.43 | 78 | 66.73 |
| ATAE-LSTM | | | 68.7 | | 77.2 | |

Table 4 – Performance of models on English datasets.

## 4 Experiments

To test Pars-ABSA dataset, 6 recent systems available for aspect-based sentiment analysis with a focus on deep learning methods were used. Table 4 compares the performance of these

models on English datasets based on f1 score macro and accuracy metrics. These methods are:

- AOA (Huang et al., 2018): An attention over attention neural network which captures the interaction between aspects and context sentences.
- Cabasc (Liu et al., 2018): This method utilizes two attention mechanisms, sentence-level content attention mechanism which captures the important information about given aspects from a global perspective, while the context attention mechanism is responsible for simultaneously taking the order of the words and their correlations into account, by embedding them into a series of customized memories.
- RAM (Chen et al., 2017): This method makes a memory from the input and with using multiple attention

mechanism, it extracts important information from memory and for prediction it uses a combination of the extracted features of different attentions non-linearly.
- IAN (Dehong et al., 2017): It learns the attentions inside the document and targets interactively, and originates the representations for targets and the document separately.
- ATAE-LSTM (Wang et al., 2016): It uses attention mechanism along with Long Short-Term Memory Network.
- TD-LSTM (Tang et al., 2016): It uses two long short-term memory networks in forward and backward directions to extract important information before and after the target and the last step of the hidden states for prediction.

For word embedding a Word2Vec[2] model was trained on all of the comments scraped from Digikala website. The result of 6 models on Pars-ABSA is given in Table 5.

Reviewing the results achieved by models, their performance on Pars-ABSA was quite impressing.

| Model | Accuracy | F1 Score |
|---|---|---|
| AOA | 79.15 | 77.13 |
| Cabasc | 68.10 | 61.90 |
| RAM | 77.35 | 74.96 |
| IAN | 76.90 | 75.09 |
| TD-LSTM | **85.54** | **84.40** |
| ATAE-LSTM | 73.80 | 72.25 |

Table 5 – Performance of models on Pars-ABSA.

But among the models, the result of TD-LSTM (Tang et al., 2016) was quite surprising. Because the other models were proposed after TD-LSTM (Tang et al., 2016) and their performances on English datasets were better than TD-LSTM (Tang et al., 2016), so it was expected from them to perform better on Pars-ABSA dataset too. The authors of RAM (Chen et al., 2017) claimed that their model is language insensitive, which mean it can perform on all languages and, compared to TD-LSTM (Tang et al., 2016) which might lose feature if the opinion word is far from the target, they employed the recurrent attention to solving this problem. But by comparing results, it's obvious that TD-LSTM (Tang et al., 2016) outperforms their method for Persian. Pars-ABSA dataset is available on a public repository[3].

## 5 Conclusion and Future Works

In this paper, Pars-ABSA, a Persian aspect-based sentiment analysis dataset gathered from Digikala website was presented also, the method of collecting and annotating the dataset and properties and statistics of the dataset was provided. Some of recent models in aspect-based sentiment analysis were used as a baseline and, their performances were compared on the dataset.

As future plans, we aim to extend Pars-ABSA with more reviews to include different domains such as restaurants and hotels. Due to the reviews and the documents posted on social medias are mostly in informal writing and the structure of sentences and even forms of the words are different from the formal writing, we are working on more advanced approaches to provide adequate tools for processing the data and a model that works properly for Persian language and the dataset.

---

[2] https://code.google.com/p/word2vec/

[3] https://github.com/Titowak/Pars-ABSA